\pdfoutput=1

\documentclass[11pt]{article}

\usepackage[]{acl}

\usepackage{times}
\usepackage{latexsym}

\usepackage[T1]{fontenc}

\usepackage[utf8]{inputenc}

\usepackage{microtype}

\usepackage{graphicx}
\usepackage{csquotes}

\title{
Divide (Text) and Conquer (Sentiment):
\\
Improved Sentiment Classification by Constituent Conflict Resolution
}

\author{Jan Kościałkowski \\
  Relativity \\
  \texttt{jan.koscialkowski@gmail.com} \\
  \And
  \And
  Paweł Marcinkowski \\
  Allegro Pay \\  
  \texttt{marcinkowski.paw@gmail.com}
}

\begin{document}
\maketitle
\begin{abstract}
Sentiment classification, a complex task in natural language processing, becomes even more challenging when analyzing passages with multiple conflicting tones. Typically, longer passages exacerbate this issue, leading to decreased model performance. The aim of this paper is to introduce novel methodologies for isolating conflicting sentiments and aggregating them to effectively predict the overall sentiment of such passages. One of the aggregation strategies involves a Multi-Layer Perceptron (MLP) model which outperforms baseline models across various datasets, including Amazon, Twitter, and SST while costing $\sim$1/100 of what fine-tuning the baseline would take.
\end{abstract}

\section{Introduction}

Sentiment classification is a difficult problem in its own right, and obscuring it by introducing multiple conflicting tones to the analysed passage does not make it any easier \cite{octa}. Take, for example, the following review.

\begin{quote}
    The staff were great, only negative was the noise, it was hard to have a conversation, and yes, we will return.
\end{quote}

In general, the overall sentiment of this passage is positive. However, a decent (not SOTA), e.g. BERT-based sentiment classifier would concentrate on the higher-represented \textbf{noise} aspect and label it as negative.

\begin{figure}[h]
    \centering
    \includegraphics[width=1\linewidth]{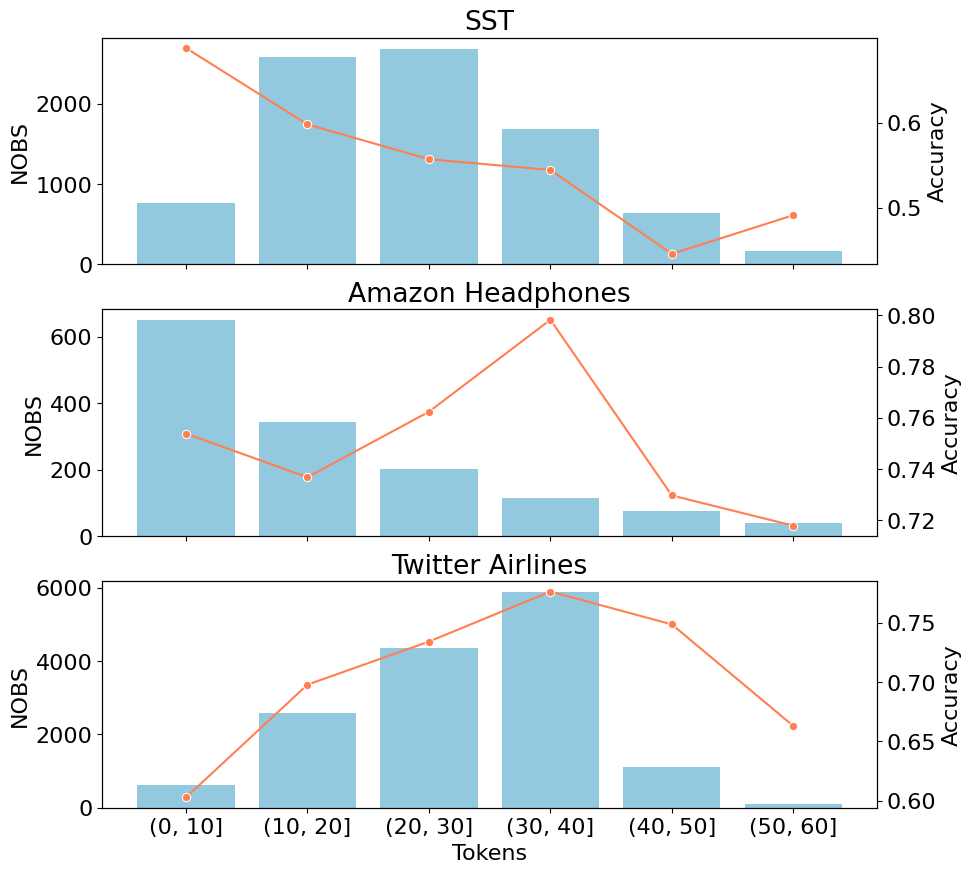}
    \caption{Model accuracy vs passage length in tokens.}
    \label{fig:perf_token}
\end{figure}

Such situations will naturally arise for longer passages. Figure \ref{fig:perf_token} depicts how with increasing passage lengths, a model's performance will inevitably start to degrade past a certain point.

In order to tackle these issues, we present several novel approaches to isolating the conflicting sentiments and later aggregating them into successful prediction of the overall sentiment for the passage at hand.

\begin{enumerate}
    \item
        Generate sentiment subpredictions
        \begin{itemize}
            \item split into sentences and generate predictions for each sentence
            \item detect aspects (using the ABSA framework) and generate predictions for each aspect
        \end{itemize}
    \item 
        Aggregate subpredictions
        \begin{itemize}
            \item Average
            \item Multilayer Perceptron on top of a feature vector of aggregates
        \end{itemize}
\end{enumerate}
    
The MLP approach is either on par or outperforms the baselines across all test datasets. If the baseline has not been fine-tuned to the specific distribution of passages, the difference in accuracy can be as big as 20 pp. Furthermore, the uplift is delivered at a cost orders of magnitude smaller than for standard fine-tuning.

Finally, as the MLP approach yields better results than the baseline, the trained model can be thought of as an approximation/proxy for patterns a more complex model would learn. Analysing its behaviour, simplified due to its constrained nature, could shed light on the inner workings of the standard sentiment classification models. Furthermore, it could be fine-tuned using Interchange Intervention Training \cite{iit}.

\section{Prior Literature}

Sentiment analysis is a well-established field which over the past years has continuously been proving itself useful as a source of well-defined and quantifiable problems to fuel the current AI boom \cite{thumbs} \cite{sst} \cite{sa_example}.

Two approaches seem to contend for the top spot in terms of sentiment classification accuracy, as found by a recent metaanalysis \cite{sota}. On the one end of the spectrum there are transformer models, mostly BERT scions \cite{bert}, fine-tuned to the specific sentiment analysis problem, or even to a specific dataset. On the other end, there are general-purpose Large Language Models (e.g. OpenAI's GPT-4 \cite{gpt4}) which achieve SOTA results using zero- or few-shot prompting. Binary sentiment classification seems to be an almost-solved problem with the best models approaching 100\% accuracy across a wide range of datasets, so we decided to concentrate on the ternary case.

One of the key components of our proposed solution is sentence and clause disambiguation. Surprisingly, the current SOTA for sentences seems to leverage a set of rules rather than an ML model. PySBD fares particularly well when the processed text contains e.g. abbreviations, numbers, and URLs \cite{pysbd}. Clauses are trickier and this area seems underexplored with very few examples of successful solutions like ClausIE \cite{Corro2013ClausIECO}.

Another way to search for atomic pieces of a passage with an unambiguous sentiment is to concentrate on \textit{aspects} rather than contiguous pieces of text. This is where Aspect-Based Sentiment Analysis (ABSA) comes into place \cite{absa}. The idea behind it is to first detect aspects pertaining to the analyzed text and then predict their sentiment in the context of the whole initial passage. There exist multiple tools supporting this approach, like \href{https://huggingface.co/docs/setfit/how_to/absa}{SetFitABSA} based on \cite{setfit}.

\section{Data}\label{sec:data}

Our selection of datasets was driven by two main motivations: to encompass a diverse range of domains where sentiment analysis can be applied, and to ensure that the datasets pose a significant challenge even for the most advanced models, as reported in recent literature.

First, predicting sentiment for the constituents of a passage immediately brings SST \cite{sst} to mind. We utilize the full-sentence-only 5-class version available on \href{https://huggingface.co/datasets/SetFit/sst5}{HuggingFace} and reduce the labels to 3 classes.

However, SST has been constructed solely from Rotten Tomatoes reviews, and thus may not be representative of all potential sentiment analysis use cases. To address this limitation, we will incorporate two additional datasets. We will enrich the review space by including a relatively recent dataset containing headphones reviews from Amazon \cite{amazon_headphones}. Additionally, we will employ a dataset consisting of X/Twitter posts about airlines, first introduced in \cite{siebert}. This source has proven notoriously difficult even for the best current models \cite{sota}, and may therefore help us to better identify the shortcomings of our approach.

For model training, the Amazon and Twitter datasets were split into train, validation and test sets using 70\%, 10\% and 20\% proportions respectively. For SST, the split maintained on Huggingface was preserved.

\section{Model}

Our approach relies on 3 substrates: a sentiment classifier, a constituent extraction heuristic and an aggregation strategy. The following subsections describe these in more detail.

\subsection{Sentiment classifiers}

When selecting models, we aimed for decent ternary sentiment classifiers, but not ones approaching SOTA so that there is some room for improvement. 

\begin{itemize}
    
    \item 
        Off-the-shelf 3-class sentiment classifier based on fine-tuned RoBERTa, introduced in \cite{hartmann2021}, available on \href{https://huggingface.co/j-hartmann/sentiment-roberta-large-english-3-classes}{HuggingFace}, later referred to as RoBERTa,
    \item
        A polarity model fine-tuned using the SetFit approach \cite{setfit}, available on \href{https://huggingface.co/tomaarsen/setfit-absa-bge-small-en-v1.5-restaurants-polarity}{HuggingFace}, later referred to as Polarity.
\end{itemize}

\subsection{Constituent extraction}
We follow two approaches to detecting constituents of the passage and generating subpredictions for them.

\begin{itemize}
    \item Rule-based: PySBD and ClauCy
    \begin{enumerate}
        \item \href{https://github.com/nipunsadvilkar/pySBD}{PySBD} - a rule-based sentence boundary detection module
        \item \href{https://github.com/mmxgn/spacy-clausie}{ClauCy} - a Python + SpaCy implementation of ClausIE. \label{clausie}
    \end{enumerate} 
    \item Aspect-based: \href{https://huggingface.co/docs/setfit/en/how_to/absa}{SetFitABSA} - fine-tuned Sentence Transformers with classification heads for aspect detection and per-aspect polarity classification which detects and assesses aspects found in a text.
\end{itemize}

Take the following review as an example.

\begin{displayquote}
    \small
        \textit{    
            If you are the type of person who does not like to fumble around with Bluetooth on windows laptops and already have a good pair of headphones and earphones and need these for taking online classes or meetings or just watching youtube, they are totally alright for that purpose.The build quality is very nice. Will survive some abuse. Have braided cable. The plastic quality is also excellent no sharp edges.I have compared these to apple's earpods. earpods still sound better than these. Don't buy these for music purposes because you'll be heavily disappointed. They sound garbage, but for anything speech related, these are fine. As I said, if you want to take a quick meeting or just for an online class or watch youtube.
    }
\end{displayquote}

The first approach yields the following clause splits and associated predictions. Some of these do not make much sense which foreshadows the problems we encountered with this approach.

\begin{table}[h!]
\small
\centering
\begin{tabular}{l|l}
\textbf{Clause} & \textbf{Values} \\ \hline
you are the type & [0.0003, 0.9983, 0.0013] \\ \hline
who like needed & [0.0029, 0.9230, 0.0741] \\ \hline
they are alright & [0.0031, 0.9925, 0.0044] \\ \hline
The build quality is very nice & [0.0002, 0.0002, 0.9996] \\ \hline
Will survive some abuse & [0.9785, 0.0207, 0.0008] \\ \hline
Have braided cable & [0.0004, 0.9984, 0.0011] \\ \hline
The plastic quality is excellent & [0.0003, 0.0002, 0.9995] \\ \hline
I have compared these & [0.0003, 0.9989, 0.0008] \\ \hline
earpods sounded still & [0.0087, 0.9903, 0.0009] \\ \hline
you be heavily disappointed & [0.9992, 0.0005, 0.0003] \\ \hline
They sounded garbage & [0.9992, 0.0005, 0.0003] \\ \hline
these are fine & [0.0002, 0.0002, 0.9996] \\ \hline
I said & [0.0005, 0.9990, 0.0005] \\ \hline
you wanted watched & [0.0008, 0.9987, 0.0005] \\
\end{tabular}
\caption{Text with associated values}
\end{table}

ABSA detects 18 aspects and classifies them as follows
\begin{table}[h!]
\small
\centering
\begin{tabular}{l|l}
\textbf{Text} & \textbf{Values} \\ \hline
Bluetooth & [0.0998, 0.0989, 0.7652] \\ \hline
windows laptops & [0.0687, 0.0828, 0.8213] \\ \hline
headphones & [0.0322, 0.0439, 0.9073] \\ \hline
earphones & [0.0662, 0.0697, 0.8383] \\ \hline
classes & [0.0626, 0.0744, 0.8361] \\ \hline
meetings & [0.0859, 0.0877, 0.7929] \\ \hline
youtube & [0.0671, 0.0649, 0.8423] \\ \hline
build quality & [0.0704, 0.0719, 0.8269] \\ \hline
abuse & [0.1139, 0.1232, 0.7162] \\ \hline
cable & [0.1175, 0.1453, 0.6917] \\ \hline
quality & [0.0274, 0.0318, 0.9267] \\ \hline
apple & [0.1237, 0.1055, 0.7255] \\ \hline
earpods & [0.1342, 0.1149, 0.7058] \\ \hline
earpods & [0.0652, 0.0675, 0.8400] \\ \hline
garbage & [0.2233, 0.1548, 0.5620] \\ \hline
speech & [0.0715, 0.0805, 0.8196] \\ \hline
meeting & [0.0968, 0.1040, 0.7632] \\ \hline
class & [0.0945, 0.0985, 0.7708] \\ \hline
youtube & [0.1120, 0.1219, 0.7243] \\
\end{tabular}
\caption{Text with associated values}
\end{table}

\subsection{Final model architecture} \label{final_model_archi}

Both of the approaches from the previous subsection yield a $N \times 3$    matrix which can be aggregated down to a 3-dimensional vector to generate the final prediction. This was done using one of the following strategies.

\begin{itemize}
    \item Average - calculate per-class mean over the column axis - used with aspect-based constituent extraction.
    \item Average WithOut Neutral sub-sentences (AWON) - exclude constituents with neutral class score > 0.9 and calculate per-class mean over the column axis - used with rule-based constituent extraction.
    \item Multilayer perceptron (MLP) - trained using feature vectors consisting of summary statistics for each class and the number of constituents.
\end{itemize}

For the MLP, the feature vector consisted of the following characteristics calculated for each class over the column axis:
\begin{itemize}
    \item mean,
    \item minimum,
    \item maximum,
    \item standard deviation,
    \item range,
    \item the number of instances when the given class had the highest score
\end{itemize}
and the number of detected constituents, so 19 features total. The model used the standard \href{https://scikit-learn.org/stable/modules/generated/sklearn.neural_network.MLPClassifier.html}{scikit-learn} implementation and had 1 hidden layer.

\section{Methods}

\subsection{Performance Metrics}

To assess the performance of our models, we utilize the following metrics:
\begin{itemize}
    \item \textbf{Accuracy}: This metric measures the proportion of correctly classified samples out of the total.
    \item \textbf{Macro Average F1-Score}: This metric evaluates the balance between precision and recall, and is calculated by taking the average F1-Score for each class, treating them equally regardless of their frequencies.
\end{itemize}

\subsection{Dimensions of Analysis}

Our analysis is conducted across several dimensions to ensure comprehensive evaluation:
\begin{itemize}
    \item \textbf{Binned Passage Length}: We categorize passages into bins based on their length in tokens (as returned by the RoBERTa tokenizer) to determine the models' performance across different passage sizes.
    \item \textbf{Datasets}: We analyze the performance for each dataset independently to account for the variability in data source and structure.
    \item \textbf{Splitting Methods}: We evaluate the effectiveness of different constituent extraction methods used to segment the passages into clauses or aspects.
    \item \textbf{Aggregation Strategies}: We compare various aggregation models used to combine base model scores of the constituents to derive the final sentiment classification.
\end{itemize}

\subsection{Optimization and Hyperparameters}

For the MLP model, the following grid of hyperparameters was considered.

\begin{itemize}
    \item Hidden layer size: \\ $[16, 32, 64, 128, 256]$
    \item Early stopping tolerance: \\ $[10^{-2}, 10^{-3}, 10^{-4}, 10^{-5}, 10^{-6}]$
    \item Epochs without improvement before early stopping: \\ $[10, 20, 30, 40, 50]$
\end{itemize}

It was fully searched using validation set accuracy as the target function. The final set of hyperparameters was:

\begin{itemize}
    \item Hidden layer size: 128
    \item Early stopping tolerance: $10^{-6}$
    \item Epochs without improvement before early stopping: 50
\end{itemize}

The full search was enabled by the fact that a single training run took several minutes on a T4 GPU (16 GB RAM). In comparison, full fine-tuning of RoBERTa on a dataset of similar size would require a larger GPU and take several hours. This translates to an approximately hundredfold reduction in training/fine-tuning cost.

\section{Results}

The implementation of all described methods and experimentation code can be found in the associated \href{https://github.com/jkoscialkowski/divide-and-conquer-sentiment}{GitHub repository}.

\subsection{Choice of passage split level}
Detailed analysis showed that ClauCy (\ref{clausie}) is not viable method of further splitting subsentences. Below example demonstrates few things:
\begin{itemize}
    \setlength\itemsep{0.01em}
    \item omitting "not" like in the first clause
    \item splitting conditionally bounded subsentences like in the first sentence of the passage
    \item in general low quality of splits - too fragmented
    \item an idea to exclude neutral constituents could be viable
    \item in general aggregation strategy could work
\end{itemize}

\textbf{Based on that we decided to focus on sentence and aspect-level splits.}\\

\subsection{Simple aggregation models}
We proceeded to compare the base models (RoBERTa and Polarity) against the Average/AWON aggregation heuristics for rule-based sentence splitting.

\begin{figure}[h]
    \centering
    \includegraphics[width=1\linewidth]{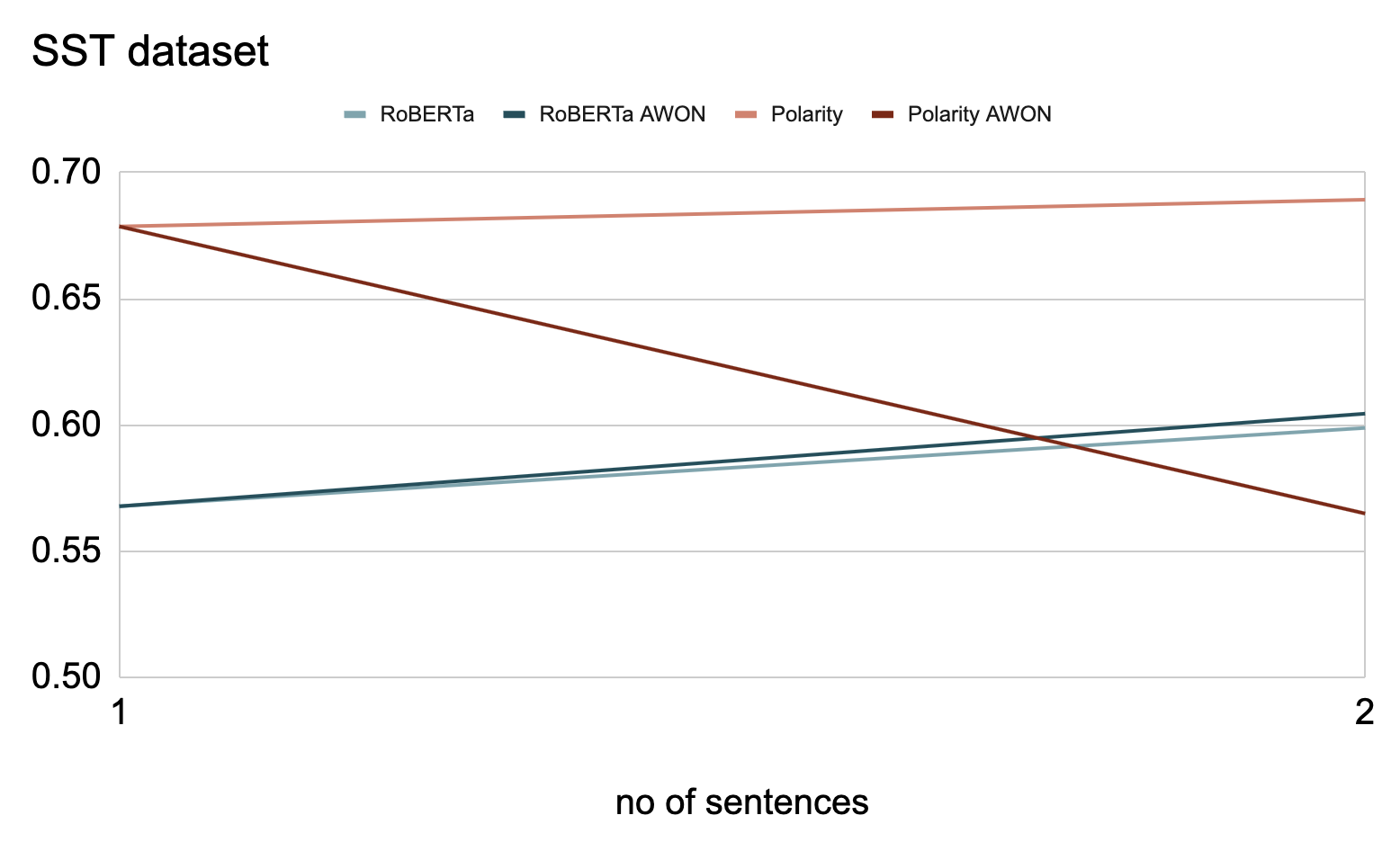}
    \caption{Accuracy of base models and Average/AWON aggregations on the SST dataset}
    \label{fig:sst_base}
\end{figure}

The SST dataset comprises brief passages, thus only a small percentage (approximately 2\%) can be segmented into individual sentences. As illustrated in Figure \ref{fig:sst_base}, the base Polarity model consistently outperforms other models, regardless of the number of sentences within a given passage.

\begin{figure}[h]
    \centering
    \includegraphics[width=1\linewidth]{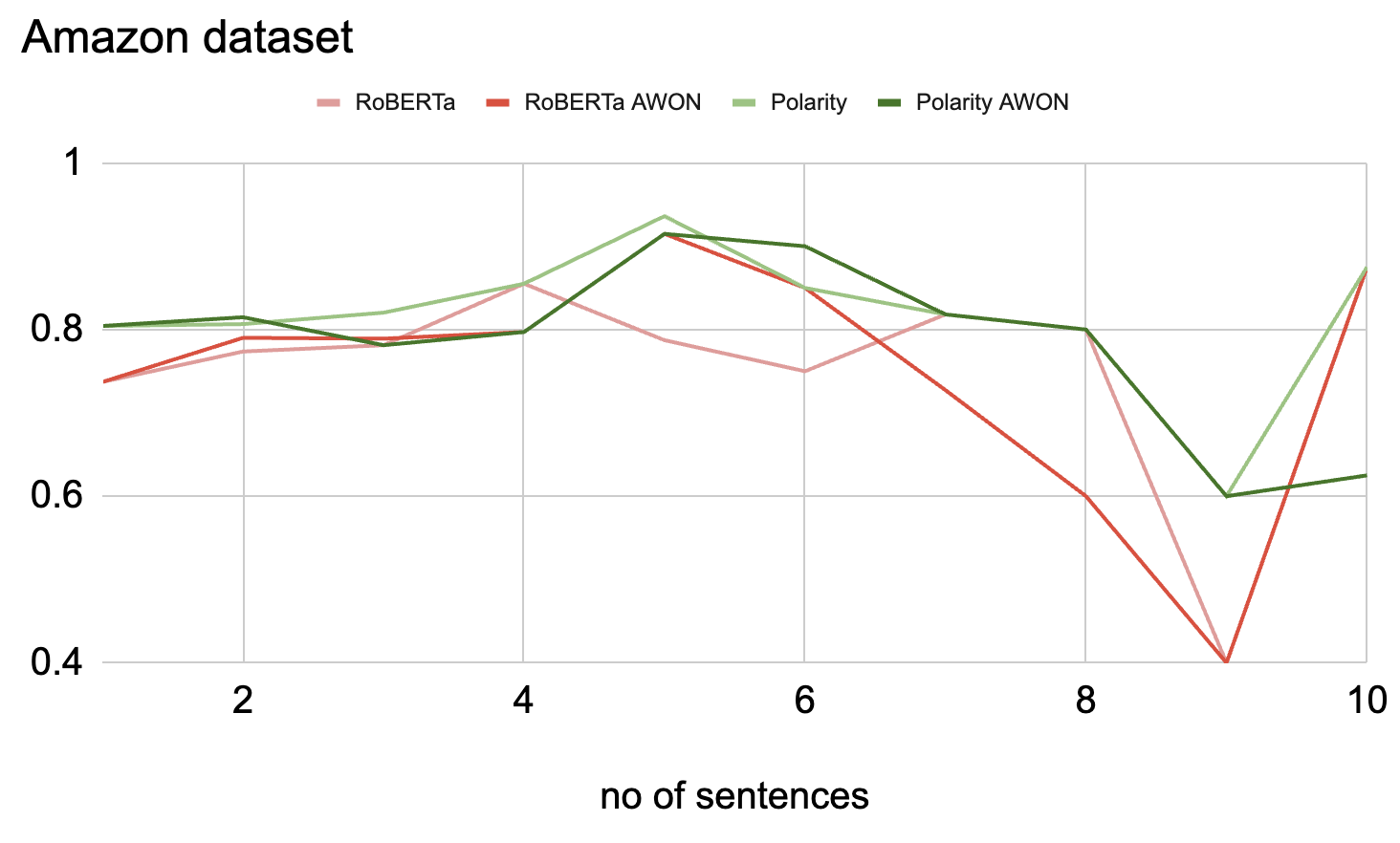}
    \caption{Accuracy of base models and Average/AWON aggregations on the Amazon dataset}
    \label{fig:amzn_base}
\end{figure}

The results obtained from the Amazon dataset are particularly noteworthy. The Base Polarity and Polarity AWON models alternate in securing the first position, with RoBERTa-based models closely following (Figure \ref{fig:amzn_base}).

\begin{figure}[h]
    \centering
    \includegraphics[width=1\linewidth]{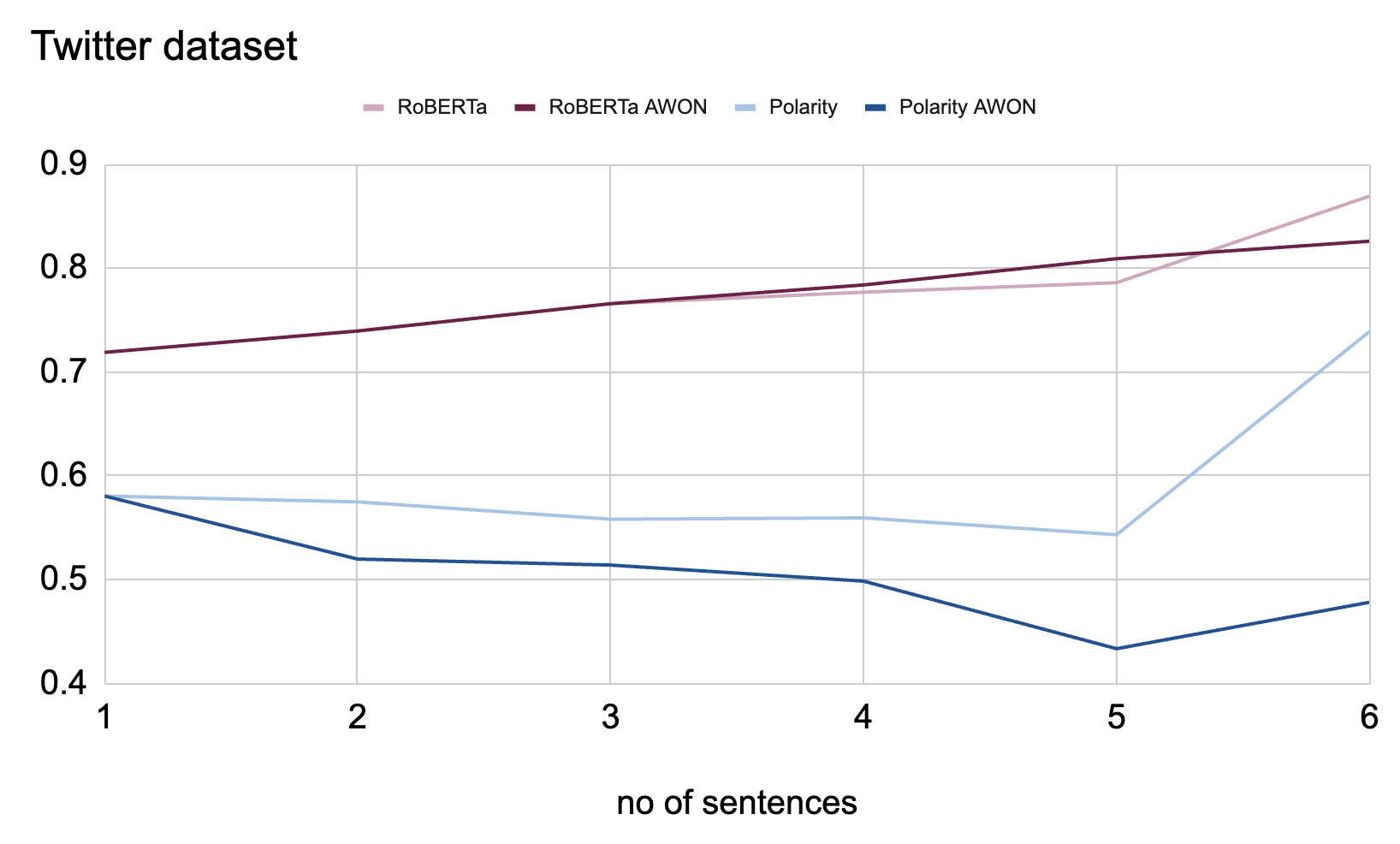}
    \caption{Accuracy of base models and Average/AWON aggregations on the Twitter dataset}
    \label{fig:twitter_base}
\end{figure}

Finally, the Twitter dataset is most effectively addressed by the RoBERTa model. The performance of both base and aggregated models is notably similar. Interestingly, the Polarity model exhibits significantly lower accuracy (by over 10 percentage points) for the base model. In addition, the performance of the aggregation models deteriorates as the number of sentences increases (see details on Figure \ref{fig:twitter_base}).

\subsection{ABSA}
Next, the aspect-based approach was tested. Here, the passage length was calculated in terms of tokens returned from the RoBERTa tokenizer and binned. We report results for both accuracy and macro averaged F1 score.

\begin{figure}[h]
    \centering
    \includegraphics[width=1\linewidth]{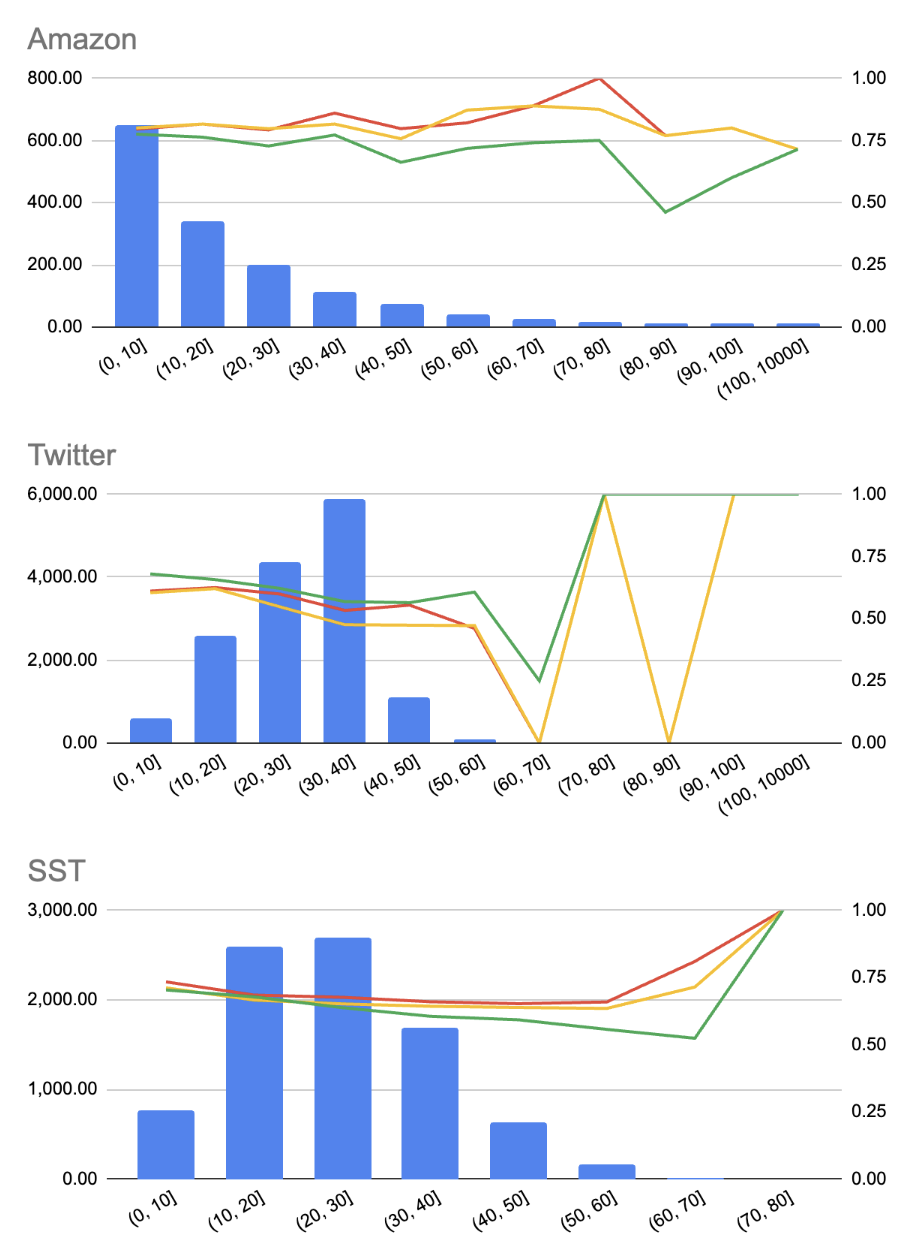}
    \caption{Accuracy of the Polarity model applied to the whole passage (red) compared to Averaged ABSA predictions using the smaller (yellow) and larger (green) models.  }
    \label{fig:absa_accuracy}
\end{figure}

Results on the Amazon and SST datasets behave similarly: there is not much difference between the base Polarity model and the ABSA approach using the smaller model, but the ABSA approach using the larger model seems to perform significantly worse. However, this observation reverses for the Twitter dataset, but the larger model's advantage is less noticeable.

\begin{figure}[h]
    \centering
    \includegraphics[width=1\linewidth]{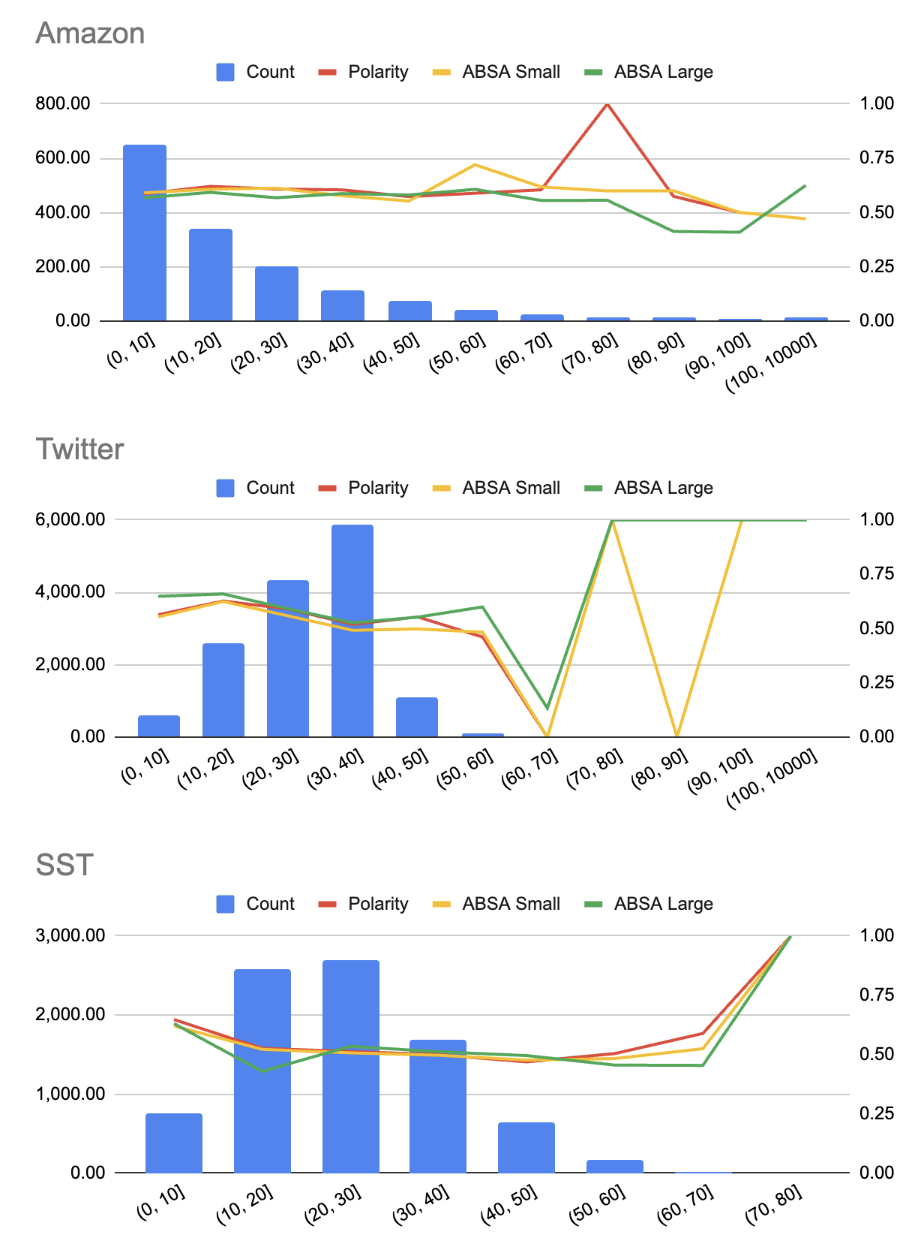}
    \caption{Macro-averaged F1 score of the Polarity model applied to the whole passage (red) compared to Averaged ABSA predictions using the smaller (yellow) and larger (green) models.}
    \label{fig:absa_macro_avg_f1}
\end{figure}

The observations for Figure \ref{fig:absa_macro_avg_f1} stay broadly the same as for Figure \ref{fig:absa_accuracy}. The differences between the smaller and larger models used in the ABSA approach seem to be somewhat less pronounced.

\subsection{MLP}
Finally, we trained an MLP model for each base model using the training sets and optimized their hyperparameters using the validation sets. Results on test sets are reported below.

\begin{table}[h!]
\centering
\begin{tabular}{|l|c|c|c|}
\hline
\textbf{} & \textbf{SST} & \textbf{Amazon} & \textbf{Twitter} \\ \hline
RoBERTa & 0.57 & 0.77 & 0.73\\ \hline
Polarity & 0.68 & 0.82 & 0.56\\ \hline
RoBERTa AWON & 0.57 & 0.78 & 0.73\\ \hline
Polarity Average & 0.68 & 0.81 & 0.52\\ \hline
MLP sentences & \textbf{0.71} & 0.80 & \textbf{0.78} \\ \hline
MLP ABSA & 0.7 & \textbf{0.83} & 0.77 \\ \hline
\end{tabular}
\caption{Comparison of Accuracy Results Across Datasets}
\end{table}

For Amazon, there seems to be no clear winner - the base Polarity model and its counterpart with averaged subpredictions challenge the MLP models. For SST, the MLP models take the lead. The Twitter results are the most interesting, because for the Polarity model \textbf{adding a shallow model on top of the base one boosted predictions by over 20 pp}; the gain is less dramatic for RoBERTa. The Polarity model was fine-tuned using restaurant reviews and RoBERTa on social media posts so, intuitively, MLP could have brought the most uplift for data of a kind the base model had not seen during its fine-tuning.

\begin{table}[h!]
\centering
\begin{tabular}{|l|c|c|c|}
\hline
\textbf{} & \textbf{SST} & \textbf{Amazon} & \textbf{Twitter} \\ \hline
RoBERTa & 0.54 & 0.63 & 0.69\\ \hline
Polarity & 0.53 & 0.59 & 0.56\\ \hline
RoBERTa AWON & 0.54 & \textbf{0.64} & 0.68\\ \hline
Polarity Average & 0.53 & 0.58 & 0.50\\ \hline
MLP sentences & \textbf{0.57} & 0.56 & \textbf{0.70} \\ \hline
MLP ABSA & 0.56 & 0.59 & 0.68 \\ \hline
\end{tabular}
\caption{Comparison of Macro Average F1-score Results Across Datasets}
\end{table}

Macro averaged F1 tells largely the same story. Amazon sees hardly any improvement from subprediction aggregation, SST sees some, and for Twitter we get massive improvements.

\section{Analysis} 

For both sentences and aspects, the tactic of simply averaging sub-predictions struggles to beat the baseline models. Only on the Twitter dataset one could observe a slight improvement (RoBERTa vs RoBERTa AWON). 

However, the MLP approach brought significant uplift in the case of SST and Twitter. It increased the accuracy by over 20 pp for a model fine-tuned on restaurant reviews (Polarity vs MLP ABSA) and by 5 pp for a model fine-tuned on social media posts (RoBERTa vs MLP sentences). These results position our approach as an efficient way to better align a pre-trained model with a given goal, at a fraction of the cost of running a fine-tuning job.

Results obtained on Macro Average F1-score confirms conclusions from the Accuracy measure. 

\section{Conclusion} 

The motivation behind this paper was to solve the issue of degrading performance for longer passages. This goal was only partially accomplished, but instead a different valuable discovery was made. Given a generic model which is to be applied to data sampled from beyond its training distribution, one can quickly augment it by using our heuristic of generating subpredictions and training a shallow MLP on top of them. Our approach delivers better results than standard fine-tuning at a $\sim$100x speedup which makes it ideal for compute-constrained regimes.

\section*{Known Project Limitations}
We have verified that performance does degrade for longer passages, but did not investigate the exact reason for that. It might be so that this phenomenon is \textit{not} caused by conflicting sentiments of the passage constituents.

The classifiers we tested were mediocre and had room for improvement that we successfully exploited. A question remains if more performant models could see any improvement after being treated with our heuristic.

The initial plan for this paper included aggregating latent representations rather than final predictions - this approach might yield 
even better results.

\section*{Authorship Statement [DONE]}

Jan Kościałkowski prepared data ingestion code, and visualizations, experimented with ABSA and did editorial work. Paweł Marcinkowski integrated sentence and clause splitter, ran performance analyses and experimented with sentence and clause aggregation. The preparation of this manuscript itself was a collaborative effort.

\bibliography{anthology,custom}

\end{document}